\RequirePackage{fix-cm}
\documentclass[twocolumn,numbook]{svjour3}          
\smartqed  
\usepackage{graphicx}
\usepackage{amssymb}
\usepackage{pifont}
\usepackage{dsfont}
\usepackage{xspace}
\usepackage{csquotes}
\usepackage{threeparttable}
\usepackage{subcaption}
\usepackage{booktabs}
\usepackage{amsmath}
\usepackage{hyperref}
\usepackage[capitalise,noabbrev]{cleveref}
\newcommand{\sat}{\texttt{SAT}\xspace}
\newcommand{\unsat}{\texttt{UNSAT}\xspace}
\newcommand{\onnx}{\texttt{ONNX}\xspace}
\newcommand{\vnnlib}{\texttt{VNN-LIB}\xspace}
\newcommand{\dnnv}{\texttt{DNNV}\xspace}
\newcommand{\cmark}{\ding{51}}%
\newcommand{\xmark}{\ding{55}}

\DeclareMathOperator{\argmax}{arg\,max}
\DeclareMathOperator{\argmin}{arg\,min}

\newcommand{\todo}[1]{\textcolor{red}{TODO: #1}}
\renewcommand{\todo}[1]{} 

\begin{document}
\emergencystretch 3em

\title{First Three Years of the International Verification of Neural Networks Competition (VNN-COMP)}


\titlerunning{VNN-COMP First Three Years}        

\author{Christopher Brix         \and
        Mark Niklas M\"uller \and
        Stanley Bak \and
        Taylor T. Johnson \and
        Changliu Liu
}


\institute{Christopher Brix \at
              RWTH Aachen University, Aachen, Germany \\
              \email{brix@cs.rwth-aachen.de}           
           \and
           Mark M\"uller \at
              ETH Zurich, Zurich, Switzerland \\
              \email{mark.mueller@inf.ethz.ch} 
           \and
           Stanley Bak \at
              Stony Brook University, Stony Brook, New York, USA \\
              \email{stanley.bak@stonybrook.edu} 
            \and
            Taylor T. Johnson \at
                Vanderbilt University, Nashville, Tennessee, USA \\
                \email{taylor.johnson@vanderbilt.edu}
            \and
            Changliu Liu \at
                Carnegie Mellon University, Pittsburgh,
Pennsylvania, USA \\
                \email{cliu6@andrew.cmu.edu}
}

\date{Received: date / Accepted: date}

\maketitle

\begin{abstract}
This paper presents a summary and meta-analysis of the first three iterations of the annual International Verification of Neural Networks Competition (VNN-COMP) held in 2020, 2021, and 2022. In the VNN-COMP, participants submit software tools that analyze whether given neural networks satisfy specifications describing their input-output behavior. These neural networks and specifications cover a variety of problem classes and tasks, corresponding to safety and robustness properties in image classification, neural control, reinforcement learning, and autonomous systems. We summarize the key processes, rules, and results, present trends observed over the last three years, and provide an outlook into possible future developments.
\end{abstract}

\section{Introduction}
Neural networks are increasingly used in safety-critical applications \cite{selfdrivingcars,acasxu}.
However, it has become apparent that they are highly susceptible to adversarial examples \cite{intriguingproperties}, i.e., minor and possibly imperceptible input perturbations can cause the output to change significantly.
As such perturbations can occur in the real world either at random or due to malicious actors, it is of utmost importance to analyze the robustness of deep learning based systems in a mathematically rigorous manner before applying them in safety-critical domains.
To this end, a wide range of methods and corresponding software tools have been developed \cite{ehlers,GehrMDTCV18,marta,reluplex}. However, with tools becoming ever more numerous and specialized, it became increasingly difficult for practitioners to decide which tool to use.

In 2020, the inaugural VNN-COMP was organized to tackle this problem and allow researchers to compare their neural network verifiers on a wide set of benchmarks.
Initially conceived as a friendly competition with little standardization, it was increasingly standardized and automated to ensure a fair comparison on cost-equivalent hardware using standardized formats for both properties and networks.

In this work, we outline this development, summarize key rules and results, describe the high-level trends observed over the last three years, and provide an outlook on possible future developments.


\section{Neural Network Verification}
We consider the neural network verification problem defined as follows: Given an input specification $\phi \subseteq \mathds{R}^{d_\text{in}}$, also called pre-condition, an output specification $\psi \subseteq \mathds{R}^{d_\text{out}}$, also called post-condition, and a neural network $N: \mathds{R}^{d_\text{in}} \mapsto \mathds{R}^{d_\text{out}}$, we aim to prove that the pre-condition implies the post-condition, i.e.,
\begin{equation}
    \forall x: x \vDash \phi \Rightarrow N(x) \vDash \psi,
    \label{eq:verification}
\end{equation}
or provide a counterexample.

Inspired by the notation common in the  SAT-solver community, we encode this problem by specifying a constraint set describing an adversarial example, i.e.,
\begin{equation}
    \exists x: x \vDash \phi \wedge N(x) \vDash \lnot \psi.
    \label{eq:negated}
\end{equation}
Therefore, we call instances where \cref{eq:negated} is satisfiable and thus the property encoded by \cref{eq:verification} does \emph{not} hold \sat and instances where \cref{eq:negated} is unsatisfiable and the property encoded by \cref{eq:verification} has been shown to hold \unsat.
Note that while it is possible to show \sat by directly searching for counter-examples using adversarial attacks  \cite{fgsm,MadryMSTV18}, those approaches are not complete, i.e., if they are not successful in finding a counter-example this does \emph{not} imply that a property holds.

\paragraph{Example Problems}
One particularly popular property is the robustness to adversarial $\ell_\infty$-norm bounded perturbations in image classification.
There, the network $N$ computes a numerical score $y \in \mathds{R}^{d_\text{out}}$ corresponding to its confidence that the input belongs to each of the $d_\text{out}$ classes for each input $x \in \mathds{R}^{d_\text{in}}$.
The final classification $c$ is then computed as $c = \argmax_i N(x)_i$.
In this setting, an adversary may want to perturb the input such that the classification changes.
Therefore, the verification intends to prove that
\begin{align*}
    \argmax_i \, & N(x')_i = t, \\
    & \forall x' \in \{x' \in \mathds{R}^{d_\text{in}} \mid \|x-x'\|_\infty \leq \epsilon\},
\end{align*}
where $t$ is the target class, $x$ is the original image and $\epsilon$ is the maximal permissible perturbation magnitude.
There, the pre-condition $\phi$ describes the inputs an attacker can choose from ($\phi = \{x' \in \mathds{R}^{d_\text{in}} \mid \|x-x'\|_\infty \leq \epsilon\}$), i.e., an $\ell_\infty$-ball of radius $\epsilon$, and the post-condition $\psi$ describes the output space corresponding to a classification to the target class $t$ ($\psi = \{y \in \mathds{R}^{d_\text{out}} \mid y_t > y_i, \forall i \neq t\}$).

When neural networks are used as controllers, more complex properties can be relevant. For example, in the ACAS Xu setting \cite{acasxu} a neural controller gives action recommendations based on the relative position and heading of the controlled and intruder aircraft. There, we want to, e.g., ensure that for inputs $\mathcal{D}$ corresponding to the intruder aircraft being straight ahead and heading our way, neither of the evasive maneuvers  "strong left" (SL) or "strong right" (SR) is considered the worst option. More formally, we want to verify that
\begin{align*}
    \argmin_i & N(x')_i \notin \{\text{SL}, \text{SR}\}, \; \forall x' \in \mathcal{D}.
\end{align*}
Here, we obtain a more complex, non-convex post-condition
\begin{align*}
    \psi = &\mathds{R}^{d_\text{out}} \setminus \\
    &\big(\{y \in \mathds{R}^{d_\text{out}} \mid y_\text{SL} < y_i, \; \forall i \notin \{\text{SL},\text{SR}\}\} \\
    &\cup \{y \in \mathds{R}^{d_\text{out}} \mid y_\text{SR} < y_i, \; \forall i \notin \{\text{SL},\text{SR}\}\} \big).
\end{align*}



\section{Competition Goals} \label{sec:goals}
VNN-COMP is organized to further the following goals.

\paragraph{Define Standards}
To enable practitioners to easily use and evaluate a range of different verification approaches and tools without substantial overhead, it is essential that all tools can process both networks and specifications in a  standardized file format.
To this end, the second iteration of the VNN-COMP established such a standard. Problem specifications (pre- and post-condition) are defined using the \vnnlib \cite{vnnlib} format and neural networks are defined using the \onnx \cite{onnx} standard.
In 2022, additionally, a standardized format for counterexamples was introduced.

\paragraph{Facilitate Verification Tool Comparison}
Every year, dozens of papers are published on neural network verification, many proposing not only new methods but also new benchmarks.
With authors potentially investing more time into tuning their method to the chosen benchmarks, a fair comparison between all these methods is difficult.
VNN-COMP facilitates such a comparison between a large number of tools on a diverse set of benchmarks, using cost-equivalent hardware, and test instances not available to participants. 
Letting participants and industry practitioners propose a wide range of interesting benchmarks, yields not only a ranking on the problems typically used in the field but also highlights which tools are particularly suitable for more specialized problems.
Further, by ensuring a standardized installation and evaluation process is in place, the comparison to a large number of state-of-the-art tools for any publication is enabled.

\paragraph{Shape Future Work Directions}
The visibility VNN-COMP lends to the problems underlying the considered benchmarks has the potential to raise their profile in the community.
As benchmarks are developed jointly by industry and academia, this constitutes a great opportunity to shape future research to be as impactful as possible.
Over the last years, benchmarks have featured ever-increasing network sizes (see \cref{table:comparison}), promoting scalability, more complex networks (including, e.g., residual \cite{resnet} and max-pooling layers \cite{maxpooling}), promoting generalizability, and more complex specifications, enabling more interesting properties to be analyzed.

\paragraph{Bring Researchers Together}
Both the rule and benchmark discussion  phase during the lead-up to the competition, as well as the in-person  presentation of results at the Workshop on Formal Methods for ML-Enabled Autonomous Systems (FoMLAS)\footnote{\url{https://fomlas2022.wixsite.com/fomlas2022}} provide participants with a great opportunity to meet fellow researches and discuss the future of the field.
Further, the tool and benchmark descriptions participants provide for the yearly report \cite{vnncomp20,vnncomp21,vnncomp22} serve as an excellent summary of state-of-the-art methods, allowing people entering the field to get a quick overview.


\section{Overview of Three Years of VNN-COMP}\label{sec:history}

In this section, we provide a high-level description of how the VNN-COMP evolved from 2020 to 2022, listing all participants and the final rankings in \cref{table:tools}.
Generally, performance is measured on a set of equally weighted \emph{benchmarks}, each consisting of a set of related \emph{instances}. Each instance consists of a trained neural network, a timeout, and input and output constraints.
Below, we group benchmarks into \emph{categories} to enable a quicker comparison between years.

\subsection{VNN-COMP 2020}\label{sec:history20}
The inaugural VNN-COMP\footnote{\url{https://sites.google.com/view/vnn20/vnncomp}} \cite{vnncomp20} was held in 2020 as a \enquote{friendly competition} with no winner. Its main goal was to provide a stepping stone for future iterations by starting the process of defining common problem settings and identifying possible avenues for standardization.

\subsubsection{Benchmarks}
Three benchmark categories were considered with only one of the eight teams participating in all of them:

\begin{itemize}
    \item Fully connected networks with ReLU activations -- two benchmarks, based on ACAS Xu and MNIST.
    \item Fully connected networks with sigmoid and tanh activation functions -- one benchmark, based on MNIST.
    \item Convolutional networks -- two benchmarks, based on MNIST and CIFAR10.
\end{itemize}

\subsubsection{Evaluation}
Teams evaluated their tools using their own hardware.
While this simplified the evaluation process, it made the reported results incomparable, due to the significant hardware differences.
The teams reported that they used between 4 and 40 CPUs and between 16 and 756 GB of RAM.

\subsection{VNN-COMP 2021}\label{sec:history21}
Based upon the insights gained in 2020, the second iteration of VNN-COMP\footnote{\url{https://sites.google.com/view/vnn2021}} was organized with a stronger focus on comparability between the participating tools \cite{vnncomp21}.

\begin{table*}[h]
\centering
\caption{Available AWS instances.}\label{table:awsinstances}
\renewcommand{\arraystretch}{1.1}
\scalebox{0.98}{
\begin{tabular}{lccccc} \toprule
         & 2021 & 2022 & vCPUs & RAM [GB] & GPU \\ 
         \midrule
         r5.12xlarge & \cmark & \xmark & 48 & 384 & \xmark \\ 
         p3.2xlarge & \cmark & \cmark & 8 & 61 & V100 GPU with 16 GB memory \\
         m5.16xlarge & \xmark & \cmark & 64 & 256 & \xmark \\
         g5.8xlarge & \xmark & \cmark & 32 & 128 & A10G GPU with 24 GB memory \\
         t2.large & \xmark & \cmark & 2 & 8 & \xmark \\
         \bottomrule
\end{tabular}
}
\end{table*}

\subsubsection{Benchmarks}
Teams were permitted to propose one benchmark with a total timeout of at most six hours split over its constituting instances.
Networks were defined in the \onnx format \cite{onnx} and problem specifications were given in the \vnnlib format \cite{vnnlib}.
%
To prevent excessive tuning to specific benchmark instances, benchmark proposers were encouraged to provide a script enabling the generation of new random instances for the final tool evaluation.
However, teams were allowed to tune their tools for each benchmark, using the initial set of benchmark instances.

In 2021, the benchmarks could be split into the following categories, with multiple teams participating in all of them:

\begin{itemize}
    \item Fully connected networks with ReLU activations -- two benchmarks, based on ACAS Xu and MNIST.
    \item Fully connected networks with sigmoid activations -- one benchmark, based on MNIST.
    \item Convolutional networks -- three benchmarks, based on CIFAR10.
    \item Networks with max-pooling layers -- one benchmark, based on MNIST.
    \item Residual networks -- one benchmark, based on CIFAR10.
    \item Large networks with sparse matrices -- one benchmark, based on database indexing.
\end{itemize}

\subsubsection{Evaluation}
To allow for comparability of results, all tools were evaluated on equal-cost hardware using  Amazon Web Services (AWS).
Each team could decide whether they wanted their tool to be evaluated on a CPU-focused r5.12xlarge or a GPU-focused p3.2xlarge instance (see \cref{table:awsinstances} for more details).
Further, instead of providing results and runtimes themselves, teams had to prepare scripts automating the installation and execution of their tools.
After the submission deadline, the organizers  installed and evaluated each tool using the provided scripts.
In many cases, this process required some debugging in a back and forth between the organizers and teams.

\paragraph{Scoring}
For every benchmark, 10 points were awarded for correctly showing the instance to be \sat/\unsat, with a 100 point penalty for incorrect results (see \cref{table:points21}).
A simple adversarial attack was used to identify \enquote{easy} \sat instances, on which the available points were reduced from 10 to 1.
If tools reported contradicting results on an instance, the ground truth was decided by a majority vote.
Bonus points were awarded to the fastest two tools on every instance (two points for the fastest and one point for the second fastest).
Runtimes differing by less than 0.2 seconds or below one second were considered equal, so multiple teams could receive the two point bonus.
To correct the notable differences in startup overhead, e.g., due to the need to acquire a GPU, it was measured as the runtime on a trivial instance and subtracted from every runtime.
\begin{table}[h]
\begin{center}
\begin{minipage}{174pt}
\caption{Points per instance in 2021. \texttt{SAT} instances were split into simple and complex based on whether a simple adversarial attack was successful.}\label{table:points21}%
\begin{tabular}{lccc} \toprule
         & \multicolumn{3}{c}{Returned Result } \\ 
         \cmidrule(lr){2-4}
         Ground Truth $\;$ &  \texttt{SAT} & \texttt{UNSAT} & Other \\ 
         \midrule
         \texttt{SAT}, simple & $+1$ & $-100$ & 0 \\
         \texttt{SAT}, complex & $+10$ & $-100$ & 0 \\
         \texttt{UNSAT} & $-100$ & $+10$ & 0\\ 
         \bottomrule
\end{tabular}
\end{minipage}
\end{center}
\end{table}
The benchmark score was computed from the points obtained as discussed above by normalizing with the maximum number of obtained points. Consequently, the tool with the most points was assigned a score of $100\%$.
The total competition score was simply the sum of the per benchmark scores, corresponding to equal weighting.

\paragraph{Results}
In 2021, 12 teams participated in the competition.
$\alpha$-$\beta$-CROWN won first place, followed by VeriNet in second, and ERAN/OVAL in third, depending on the overhead measurement and voting scheme used to determine result-correctness.
Except for VeriNet, they all used the GPU instance.

\subsection{VNN-COMP 2022}\label{sec:history22}
In the most recent iteration of VNN-COMP\footnote{\url{https://sites.google.com/view/vnn2022}} \cite{vnncomp22}, the evaluation was fully automated, allowing the number of benchmarks to be increased.

\subsubsection{Benchmarks}
In 2022, each participating team could submit or endorse up to two benchmarks, allowing industry practitioners to propose benchmarks without entering a tool.
Each benchmark had a total timeout of between three and six hours, with randomization of instances being mandatory this year.
Tool tuning was still permitted on a per benchmark level and in practice also per network using the network's statistics.

The submitted benchmarks can be grouped into the following categories:

\begin{itemize}
    \item Fully connected networks with ReLU activations -- three benchmarks, based on reinforcement tasks and MNIST.
    \item Fully connected networks in TLL format \cite{tll} -- one benchmark.
    \item Large networks with sparse matrices -- one benchmark, based on database indexing and cardinality estimation.
    \item Convolutional networks -- three benchmarks, based on CIFAR10.
    \item Residual networks -- two benchmarks, based on CIFAR10, CIFAR100, and TinyImageNet.
    \item Complex U-Net networks with average-pooling and softmax -- one benchmark based on image segmentation. 
\end{itemize}

\subsubsection{Evaluation}
Similar to the previous year, teams could choose between a range of AWS instance types (see \cref{table:awsinstances}) providing a CPU, GPU, or mixed focus.
Except for the much weaker t2.large instance, all instances were priced at around three dollars per hour.
In contrast to 2021 where organizers had to manually execute installation scripts and debug with the participants, an automated submission and testing pipeline was set up.
Teams could submit their benchmarks and tools via a web interface by specifying a git repository, commit hash, and post-installation script (enabling, e.g., the acquisition of licenses).
This triggered a new AWS instance to be spawned where all installation scripts were executed.
If the installation succeeded, the tool was automatically evaluated on a previously selected set of benchmarks before the instance was terminated again.
To enable debugging by the participants, all outputs were logged and made accessible live via the submission website, allowing them to monitor the progress.
This automation allowed each team to perform as many tests as necessary without the need to wait for feedback from the organizers.
Furthermore, teams could test on the same AWS instances used during final evaluation without having to pay for their usage, with the costs kindly covered by the SRI Lab of ETH Zurich.

\paragraph{Scoring}
Unlike during the VNN-COMP 2021, \sat instances were not divided into simple and complex for scoring purposes, leading to 10 points being awarded for all correct results (see \cref{table:points22}).
Further, instead of relying on a voting scheme to determine the ground truth in the presence of dissent among tools, the burden of proof was placed on the tool reporting \sat, requiring them to provide a concrete counter-example.
If no valid counter-example was provided, the corresponding tool was judged to be incorrect and awarded the 100 point penalty.

\begin{figure*}[h]
\centering
\centerline{\includegraphics[width=\linewidth]{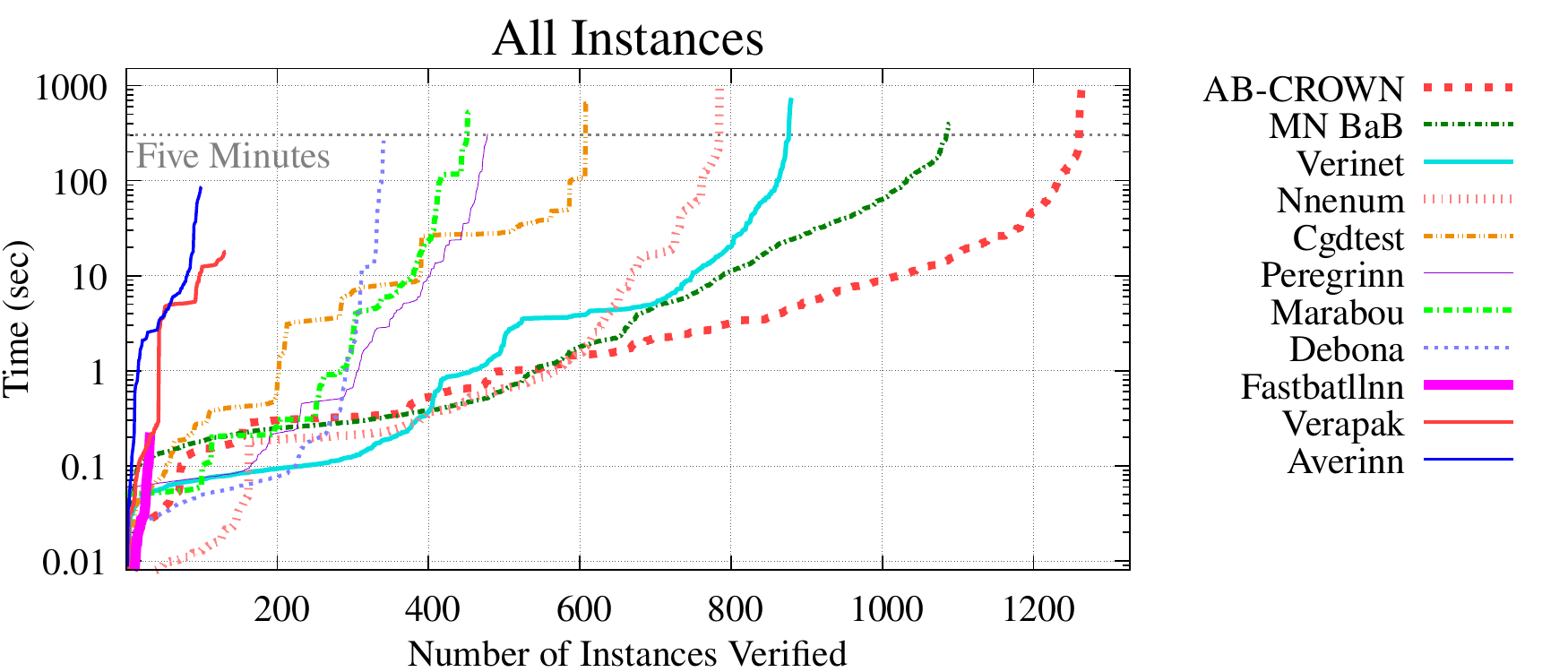}}
\caption{Cactus plot for all tools in the VNN-COMP 2022 across all benchmarks.}
\label{fig:quantPic}
\end{figure*}

\paragraph{Results}
Out of the eleven participating teams, $\alpha$-$\beta$-CROWN placed first, \textsc{MN-BaB} second, and VeriNet third.
For a comparison of all participating tools across all benchmarks, see Figure~\ref{fig:quantPic}.

\begin{table}[h]
\centering
\caption{Points per instance in 2022.}\label{table:points22}%
\begin{tabular}{lccc} \toprule
         & \multicolumn{3}{c}{Returned Result } \\ 
         \cmidrule(lr){2-4}
         Ground Truth $\;$ &  \texttt{SAT} & \texttt{UNSAT} & Other \\ 
         \midrule
         \texttt{SAT} & $+10$ & $-100$ & 0 \\
         \texttt{UNSAT} & $-100$ & $+10$ & 0\\ 
         \bottomrule
\end{tabular}
\end{table}

\section{Comparison Across the Years}
In \cref{table:tools} we list all tools participating in any iteration of the VNN-COMP and refer the interested reader to the corresponding VNN-COMP report for a short description of the tools.
In \cref{table:comparison}, we compare the scope of the competition across the last three years.
As can be seen, the number, variety, complexity, and scale of benchmarks increased with every iteration. Starting with 5 benchmarks covering simple fully connected (FC) and convolutional (Conv) networks in 2020, the 2022 competition saw 12 benchmarks including a range of complex residual and U-Net architectures with up to 140 million parameters.
Further, we believe that the increasing number of registered tools clearly shows that the interest in both the field in general and the competition in particular is growing year by year.
However, the large and increasing discrepancy between registered and submitted tools might indicate that many teams feel like they are not able to invest the significant effort required to support not only the standardized network and specification formats but also the wide variety of different benchmarks.
As tools are ranked by their total score with each benchmark providing a score of up to 100$\%$, the final ranking is biased towards tools that support all benchmarks.
While we believe that this is a valuable incentive for tool developers to develop methods that can be easily applied to new problems, it might be daunting for new teams to implement all necessary features, deterring them from participating at all.

\paragraph{Successful Trends}
While all teams started out using only CPUs in 2020, only one of the top four teams relied solely on CPUs in 2021, and all top three teams chose GPU instances in 2022. This transition enabled both the more efficient evaluation of simple bound propagation methods such as DeepPoly \cite{eran3}, CROWN \cite{ZhangWCHD18}, and IBP \cite{GowalIBP} and approximate solutions of the linear programming (LP) problems arising during verification \cite{FerrariMJV22,abcrown4,abcrown3}.
Similarly, the top two teams in 2021 and all top three teams in 2022 relied on a branch-and-bound (BaB) based approach, recursively breaking down the verification problem into easier subproblems until it becomes solvable, thus effectively enabling the use of GPUs to solve tighter mixed integer linear programming (MILP) encodings of the verification problem \cite{oval2,FerrariMJV22,abcrown4,ZhangGCP}.
Both top two teams in the most recent iteration combined this approach with additional multi-neuron \cite{FerrariMJV22} and solver-generated cutting plane constraints \cite{ZhangGCP}, first introduced by the 3rd place ERAN in 2021 \cite{eran5}.
We thus conclude that successful tools leverage hardware accelerators such as GPUs to efficiently handle tight (MI)LP encodings of the verification problem.

\begin{table*}[h]
\begin{center}
\begin{minipage}{\linewidth}
\centering
\caption{Participating tools. \todo{add urls, probably using package just on the toolname, otherwise probably too ugly, e.g., some don't have refs, but do have github, so can find them: https://github.com/formal-verification-research/VERAPAK}}%
\label{table:tools}%
\begin{tabular}{lp{5.0cm}llllll} \toprule
         Tool & Organization & \multicolumn{3}{c}{Participation, Place} & References \\
         \cmidrule{3-5}
          &  & 2020 \cite{vnncomp20} & 2021 \cite{vnncomp21} & 2022 \cite{vnncomp22} &  \\
         \midrule
         $\alpha$-$\beta$-CROWN & Carnegie Mellon, Northeastern, Columbia,
UCLA & \xmark & \cmark (1/12) & \cmark (1/11) & \cite{abcrown1,abcrown2,abcrown3} \\
         AveriNN & Kansas State University & \xmark & \xmark & \cmark (11/11) & N/A \\
         CGDTest & University of Waterloo & \xmark & \xmark & \cmark (5/11) & N/A \\
         Debona & RWTH Aachen University & \xmark & \cmark (6/12) & \cmark (8/11) & \cite{debona} \\
         DNNF & University of Virginia & \xmark & \cmark (12/12) & \xmark & \cite{dnnf} \\
         \textsc{ERAN} & ETH Zurich, UIUC & \cmark & \cmark (3/12) & \xmark & \cite{eran1,eran2,eran3,eran4,eran5,eran6} \\
         FastBatLLNN & University of California & \xmark & \xmark & \cmark (9/11) & N/A \\
         Marabou & Hebrew University of Jerusalem, Stanford University, Amazon Web Services, NRI Secure & \xmark & \cmark (5/12) & \cmark (7/11) & \cite{marabou} \\
         MIPVerify & Massachusetts Institute of Technology & \cmark & \xmark & \xmark & \cite{mipverify} \\
         \textsc{MN-BaB} & ETH Zurich & \xmark & \xmark & \cmark (2/11) & \cite{FerrariMJV22} \\
         nnenum & Stony Brook University & \cmark & \cmark (8/12) & \cmark (4/11) & \cite{nnenum} \\
         NNV & Vanderbilt University & \cmark & \cmark (9/12) & \xmark & \cite{nnv1,nnv2,nnv3,nnv4,nnv5} \\
         NV.jl & Carnegie Mellon, Northeastern & \xmark & \cmark (10/12) & \xmark & \cite{nvjl} \\
         Oval & University of Oxford & \cmark & \cmark (3/12) & \xmark & \cite{oval1,oval2,oval3} \\
         PeregriNN & University of California & \cmark & \xmark & \cmark (6/11) & \cite{peregrinn} \\
         RPM & Stanford & \xmark & \cmark (11/12) & \xmark & \cite{rpm} \\
         Venus & Imperial College London & \cmark & \cmark (7/12) & \xmark & \cite{venus1,venus2} \\
         VeraPak & Utah State University & \xmark & \xmark & \cmark (10/11) & N/A \\
         VeriNet & Imperial College London & \cmark & \cmark (2/12) & \cmark (3/11) & \cite{verinet1,verinet2} \\
         \bottomrule
\end{tabular}
\end{minipage}
\end{center}
\end{table*}

\begin{table*}[h]
\begin{center}
\begin{minipage}{\linewidth}
\centering
\caption{Comparison across years.}\label{table:comparison}%
\begin{tabular}{p{4.0cm}p{3.45cm}p{3.45cm}p{3.45cm}} \toprule
          $\;$ &  2020 & 2021 & 2022 \\ 
         \midrule
         Tools registered & N/A & 15 & 18 \\
         Tools submitted & 8 & 13 & 11 \\
         Benchmarks submitted & 5 & 8 (+1 unscored) & 12 (+1 unscored)
         \\
         Max. network depth & 8 & 18 & 27 \\
         Max. network parameters & 855,600 & 42,059,431 (sparse) & 138,356,520 \\
         Activation functions & ReLU, tanh, sigmoid & ReLU, sigmoid, MaxPool, AveragePool & ReLU, sigmoid, MaxPool \\
         Layer types & Fully Connected, Conv & Fully Connected, Conv, Residual & Fully Connected, Conv, Residual, BatchNorm \\
         Applications & Image Recognition, Control & Image Recognition, Control, Database Indexing & Image Recognition, Control, Database Indexing, Cardinality Estimation \\
         Mean \#benchmarks/tool & 3.0 (min 2, max 5) & 5.5 (min 1, max 9) & 7.3 (min 1, max 13) \\
         \bottomrule
\end{tabular}
\end{minipage}
\end{center}
\end{table*}

\section{Outlook}
Below we discuss considerations that could enable future iterations of the VNN-COMP to serve its goals and the community, discussed in \cref{sec:goals}, even better.

\subsection{Tracking Year-on-Year Progress}
While we believe VNN-COMP already provides reasonable mechanisms for comparing the tools submitted in every iteration, the changing benchmarks and tools make it hard to track the year-on-year progress of the field as a whole.
Because some tools are heavily optimized for the specific benchmarks of that year's competition, simply evaluating them on the benchmarks of previous (or future) years (even if they support them) does not yield a meaningful progress metric.
While one benchmark from the inaugural competition was included as an unscored extra benchmark in the two following iterations (\texttt{cifar2020}), only few unsolved instances remain, making it a very insensitive measure for further improvements.
While including all benchmarks from previous years in the (scored) benchmark selection would place an undue burden on participants, choosing one particularly challenging, representative, and interesting benchmark every year to be included as a (scored) extra benchmark in future iterations might be a good compromise. Additionally, a more restrictive stance on tool tuning could enable a much more representative evaluation of new tools on old benchmarks.

\subsection{Tool Tuning}
Many of the most successful tools do not employ a single verification strategy, but a whole portfolio of different modes, all coming with different hyperparameters. 
Depending on their choice, tool performance can vary significantly, making it essential for practitioners to get their choice right when applying these tools to new problems.
However, this can be highly challenging given the large number of parameters and their complex interactions, especially without in-depth knowledge of the tool.

For VNN-COMP, tuning tools was allowed explicitly on a per-benchmark basis and implicitly on a per-network basis, enabling teams to showcase the maximum performance of their tools. However, for future iterations it might be interesting to restrict tuning for some or all benchmarks to encourage authors to develop autotuning strategies, making the adaption of their tools to new problems much easier.
This could, for example, be implemented by not only generating random specifications but also random networks.

\subsection{Batch Processing}
Every VNN-COMP benchmark consists of a set of instances that, while typically related, are evaluated in isolation, with the tool being terminated in between.
Unfortunately, this means that any startup overhead such as acquiring a GPU or preprocessing the considered network is incurred for every instance. 
This is in contrast to most practical settings where a large number of input-output specifications are considered for the same network.
This discrepancy is accounted for by measuring and subtracting this overhead from each individual runtime.
However, not only is this overhead measurement process flawed and introduces noise, but it can also dominate the evaluation time for easy instances.

In future iterations, tools could be provided with a whole batch of properties at once to more closely relate to their typical application.
Further, currently, timeouts are defined per instance, making a strategy of always attempting verification until timeout optimal. 
However, in a practical setting, recognizing instances where verification is likely to fail and stopping early can significantly increase a method's throughput and thus utility. 
Switching to per benchmark timeouts for the VNN-COMP would incentivize the development of effective heuristics towards this goal.
Furthermore, tools could benefit from proof-sharing approaches \cite{FischerSDSV22}, where verified sub-problems from one instance are reused for following instances.

\subsection{Continuous Competition}
In addition to a yearly VNN-COMP, tool submissions for the most recent benchmark set could be accepted on a rolling basis, made possible by the automated submission and evaluation process introduced this year.
This would transform the competition from a yearly snapshot of the current research to a centralized repository of the state-of-the-art, updating as teams submit new methods that they publish.
However, if not implemented with great care, this would enable tools to be tuned on the evaluation instances before submission, leading to a skewed comparison.
Further, the question of funding the required cloud compute remains open.


\subsection{Soundness Evaluation}
An inherent requirement for neural network verifiers is that they are sound, i.e., they never claim a safety property holds, when in fact it does not.
However, assessing soundness is difficult as the ground truth for VNN-COMP problem instances is generally only known if it was shown to be \sat with a valid counter-example.
This is particularly problematic when no instances in a benchmark are \sat and thus returning \unsat for every instance immediately can not be demonstrated to be unsound. Requiring a certain portion of instances of every benchmark to be \sat (in expectation), could alleviate this issue.
An interesting alternative avenue to tackle this challenge is proof generation \cite{proofgeneration}. An extra category could be introduced where tools are additionally required or incentivized to provide a verifiable proof if they claim a property is \unsat.

While big soundness bugs are rare, few or none of the submitted tools are floating point sound, i.e., even tools that would be sound using exact arithmetic might become unsound due to imprecisions introduced by floating-point arithmetic. This is particularly pronounced if tools choose to use single precision computations for performance reasons. 
The sensitivity of different tools to such issues could be evaluated on a benchmark specifically designed to uncover floating point soundness issues.

\subsection{Other Competition Modes}
A dedicated falsifier category could be added to encourage teams to develop and submit stronger attacks, going beyond the standard adversarial attacks.
Further, a meta-solver category could be added to investigate whether approaches that heuristically pick from a range of methods, successful in other domains \cite{metasat}, can significantly outperform individual tools. However, it would need to be ensured that these tools provide sufficient value over individual submissions, which already combine different verification strategies.

\subsection{Promote Common Tool Development}
Parsing large and complex \vnnlib files or converting \onnx files to other common formats can be time-consuming to implement.
While many teams implemented their own tools to this end, available, open-source tools for the parsing of \vnnlib files \cite{vnnlibparser} and the optimization of \onnx files (\dnnv \cite{dnnv}) should be highlighted and their continued development encouraged.

\subsection{Remaining Challenges}
We can broadly identify four groups of challenges in neural network verification:
\begin{itemize}
    \item Verifying relatively small but only weakly regularized networks, which requires an extraordinarily precise analysis, can still be intractable with current methods.
    \item Scaling precise methods to medium-sized networks (e.g. small ResNets) and datasets (e.g. Cifar100 or TinyImageNet) with a large number of neurons is challenging, as the cost of branch-and-bound based algorithms scales exponentially with the required split depth, making branching decisions both harder and more important.
    \item Scaling verification to large networks (e.g. VGG-Net 16) and datasets (e.g. ImageNet) in the presence of dense input specifications requires particularly memory-efficient implementations due to a large number of neurons.
    \item Verification outside of the classification setting is underexplored leading to a lack of established approaches for, e.g., image segmentation or object detection.
\end{itemize}
Orthogonally, the training of certifiably robust networks remains an open problem. Despite significant progress over recent years \cite{BalunovicV20,GowalIBP,MirmanGV18,MullerSABR,PalmaIBPR,ShiWZYH21,ZhangCXGSLBH20}, networks trained specifically to exhibit provable robustness guarantees still suffer from severely degraded standard accuracy. Therefore, most benchmarks considered in the VNN-COMP are based on networks trained without consideration for later certification.
More broadly in the community, readers may also be interested in the International Competition on Verifying Continuous and Hybrid Systems (ARCH-COMP)\footnote{\url{https://cps-vo.org/group/ARCH/FriendlyCompetition}} category on Artificial Intelligence and Neural Network Control Systems (AINNCS), which has been held annually since 2019~\cite{lopez2019archcomp_ainncs,lopez2022archcomp_ainncs}, and considers neural network verification in closed-loop systems.

\section{Advice for Participants}
In this section, we provide some guidance for teams that are interested in the VNN-COMP but have not participated yet. Note that these are neither rules nor requirements.

\subsection{For Benchmark Authors}
The VNN-COMP intends to highlight areas where neural network verification can be successfully applied and to showcase interesting differences between the participating tools.
Thus, ideally, tasks are not so hard that none of the instances can be solved by any participant but also not so easy that every tool can solve all of them.
For benchmarks related to real-world applications, we recommend including a detailed description of the background, to highlight the benchmark's relevance and the characteristics of the verification problem, e.g. sparseness of the input or some network layers.
Further questions and requests for modifications should be expected while tool authors work on supporting the proposed benchmark.

\subsection{For Tool Authors}
We recommend teams reference past benchmarks to test their tool before the new benchmarks are submitted.
Given the ever-increasing diversity of submitted benchmarks, it may not be feasible to support all benchmarks from the get-go. If this is the case, we recommend focusing on the fully connected and convolutional ReLU networks, which in the past have covered a wide range of benchmarks, while minimizing implementation effort.
Some operations, e.g., max-pooling can also be simplified to multiple ReLU layers using tools such as \dnnv \cite{dnnv}.
Further, we recommend extensive testing against adversarial attacks to minimize the chance for soundness errors.
For tools that are designed for very specific problems, we also want to encourage authors to submit a relevant benchmark highlighting this specialization.
Finally, we recommend reading publications associated with the well-performing tools (see \cref{table:tools}) to gain a better understanding of the techniques used by successful teams.


\section{Conclusions}
In this report, we summarize the main processes and results of the three VNN-COMP held so far from 2020 to 2022.
We highlight the growing interest in the field, expressed in an increasing number of registered teams and considered benchmarks, including some submitted by industry.
Further, we observe that every year, the size and complexity not only of the considered networks but also specifications grew, driving and exemplifying progress in the field.
Finally, we highlight the increase in accessibility of verification methods resulting from the standardized input and output formats and the automated installation and evaluation process required for participation in VNN-COMP.


\begin{acknowledgements}
This material is based upon work supported by the Air Force Office of Scientific Research and the Office of Naval Research under award numbers FA9550-19-1-0288, FA9550-21-1-0121, FA9550-22-1-0019, FA9550-22-1-0450, and N00014-22-1-2156, as well the Defense Advanced Research Projects Agency (DARPA) Assured Autonomy program through contract number FA8750-18-C-0089, and the National Science Foundation (NSF) under grants 1911017, 2028001, 2220401, and 2220426. Any opinions, findings, and conclusions or recommendations expressed in this material are those of the author(s) and do not necessarily reflect the views of the United States Air Force, the United States Navy, DARPA, nor NSF.
\end{acknowledgements}

\bibliographystyle{spmpsci}
\bibliography{sn-bibliography}

\begin{thebibliography}{10}
\providecommand{\url}[1]{{#1}}
\providecommand{\urlprefix}{URL }
\expandafter\ifx\csname urlstyle\endcsname\relax
  \providecommand{\doi}[1]{DOI~\discretionary{}{}{}#1}\else
  \providecommand{\doi}{DOI~\discretionary{}{}{}\begingroup
  \urlstyle{rm}\Url}\fi

\bibitem{vnnlibparser}
{Simple Adversarial Generator}.
\newblock \url{https://github.com/stanleybak/simple_adversarial_generator}.
\newblock Accessed: 2022-09-13

\bibitem{vnncomp20}
{VNN-COMP2020} report.
\newblock \url{https://www.overleaf.com/project/5f0c85e8d15dc10001749fa9}.
\newblock Accessed: 2022-08-28

\bibitem{onnx}
Bai, J., Lu, F., Zhang, K., et~al.: Onnx: Open neural network exchange.
\newblock \url{https://github.com/onnx/onnx} (2019)

\bibitem{nnenum}
Bak, S.: Execution-guided overapproximation (ego) for improving scalability of
  neural network verification (2020)

\bibitem{vnncomp21}
Bak, S., Liu, C., Johnson, T.: The second international verification of neural
  networks competition ({VNN-COMP 2021}): Summary and results (2021).
\newblock \doi{10.48550/ARXIV.2109.00498}.
\newblock \urlprefix\url{https://arxiv.org/abs/2109.00498}

\bibitem{BalunovicV20}
Balunovic, M., Vechev, M.T.: Adversarial training and provable defenses:
  Bridging the gap.
\newblock In: 8th International Conference on Learning Representations, {ICLR}
  2020, Addis Ababa, Ethiopia, April 26-30, 2020. OpenReview.net (2020).
\newblock \urlprefix\url{https://openreview.net/forum?id=SJxSDxrKDr}

\bibitem{selfdrivingcars}
Bojarski, M., Del~Testa, D., Dworakowski, D., Firner, B., Flepp, B., Goyal, P.,
  Jackel, L.D., Monfort, M., Muller, U., Zhang, J., Zhang, X., Zhao, J., Zieba,
  K.: End to end learning for self-driving cars (2016).
\newblock \doi{10.48550/ARXIV.1604.07316}.
\newblock \urlprefix\url{https://arxiv.org/abs/1604.07316}

\bibitem{venus1}
Botoeva, E., Kouvaros, P., Kronqvist, J., Lomuscio, A., Misener, R.: Efficient
  verification of neural networks via dependency analysis.
\newblock In: Proceedings of the 34th AAAI Conference on Artificial
  Intelligence (AAAI20). {AAAI} Press (2020)

\bibitem{debona}
Brix, C., Noll, T.: Debona: Decoupled boundary network analysis for tighter
  bounds and faster adversarial robustness proofs.
\newblock CoRR \textbf{abs/2006.09040} (2020).
\newblock \urlprefix\url{https://arxiv.org/abs/2006.09040}

\bibitem{oval1}
Bunel, R., De~Palma, A., Desmaison, A., Dvijotham, K., Kohli, P., Torr, P.H.,
  Kumar, M.P.: Lagrangian decomposition for neural network verification.
\newblock Conference on Uncertainty in Artificial Intelligence  (2020)

\bibitem{oval2}
Bunel, R., Lu, J., Turkaslan, I., Kohli, P., Torr, P., Kumar, M.P.: Branch and
  bound for piecewise linear neural network verification.
\newblock Journal of Machine Learning Research \textbf{21}(2020) (2020)

\bibitem{ehlers}
Ehlers, R.: Formal {Verification} of {Piece}-{Wise} {Linear} {Feed}-{Forward}
  {Neural} {Networks}.
\newblock In: International {Symposium} on {Automated} {Technology} for
  {Verification} and {Analysis}, pp. 269--286 (2017).
\newblock \doi{10/gh25vg}

\bibitem{tll}
Ferlez, J., Shoukry, Y.: Aren: Assured relu nn architecture for model
  predictive control of lti systems.
\newblock In: Proceedings of the 23rd International Conference on Hybrid
  Systems: Computation and Control, HSCC '20. Association for Computing
  Machinery, New York, NY, USA (2020).
\newblock \doi{10.1145/3365365.3382213}.
\newblock \urlprefix\url{https://doi.org/10.1145/3365365.3382213}

\bibitem{FerrariMJV22}
Ferrari, C., M{\"{u}}ller, M.N., Jovanovic, N., Vechev, M.T.: Complete
  verification via multi-neuron relaxation guided branch-and-bound.
\newblock In: The Tenth International Conference on Learning Representations,
  {ICLR} 2022, Virtual Event, April 25-29, 2022. OpenReview.net (2022).
\newblock \urlprefix\url{https://openreview.net/forum?id=l\_amHf1oaK}

\bibitem{FischerSDSV22}
Fischer, M., Sprecher, C., Dimitrov, D.I., Singh, G., Vechev, M.T.: Shared
  certificates for neural network verification.
\newblock In: S.~Shoham, Y.~Vizel (eds.) Computer Aided Verification - 34th
  International Conference, {CAV} 2022, Haifa, Israel, August 7-10, 2022,
  Proceedings, Part {I}, \emph{Lecture Notes in Computer Science}, vol. 13371,
  pp. 127--148. Springer (2022).
\newblock \doi{10.1007/978-3-031-13185-1\_7}.
\newblock \urlprefix\url{https://doi.org/10.1007/978-3-031-13185-1\_7}

\bibitem{GehrMDTCV18}
Gehr, T., Mirman, M., Drachsler{-}Cohen, D., Tsankov, P., Chaudhuri, S.,
  Vechev, M.T.: {AI2:} safety and robustness certification of neural networks
  with abstract interpretation.
\newblock In: 2018 {IEEE} Symposium on Security and Privacy, {SP} 2018,
  Proceedings, 21-23 May 2018, San Francisco, California, {USA}, pp. 3--18.
  {IEEE} Computer Society (2018).
\newblock \doi{10.1109/SP.2018.00058}.
\newblock \urlprefix\url{https://doi.org/10.1109/SP.2018.00058}

\bibitem{fgsm}
Goodfellow, I.J., Shlens, J., Szegedy, C.: Explaining and harnessing
  adversarial examples.
\newblock In: Y.~Bengio, Y.~LeCun (eds.) 3rd International Conference on
  Learning Representations, {ICLR} 2015, San Diego, CA, USA, May 7-9, 2015,
  Conference Track Proceedings (2015).
\newblock \urlprefix\url{http://arxiv.org/abs/1412.6572}

\bibitem{GowalIBP}
Gowal, S., Dvijotham, K., Stanforth, R., Bunel, R., Qin, C., Uesato, J.,
  Arandjelovic, R., Mann, T.A., Kohli, P.: On the effectiveness of interval
  bound propagation for training verifiably robust models.
\newblock CoRR \textbf{abs/1810.12715} (2018).
\newblock \urlprefix\url{http://arxiv.org/abs/1810.12715}

\bibitem{resnet}
He, K., Zhang, X., Ren, S., Sun, J.: Deep residual learning for image
  recognition.
\newblock In: 2016 IEEE Conference on Computer Vision and Pattern Recognition
  (CVPR), pp. 770--778 (2016).
\newblock \doi{10.1109/CVPR.2016.90}

\bibitem{verinet1}
Henriksen, P., Lomuscio, A.: Efficient neural network verification via adaptive
  refinement and adversarial search.
\newblock In: Proceedings of the 24th European Conference on Artificial
  Intelligence (ECAI20) (2020)

\bibitem{verinet2}
Henriksen, P., Lomuscio, A.: Deepsplit: An efficient splitting method for
  neural network verification via indirect effect analysis.
\newblock In: Proceedings of the 30th International Joint Conference on
  Artificial Intelligence (IJCAI21) (To appear, August 2021)

\bibitem{marta}
Huang, X., Kwiatkowska, M., Wang, S., Wu, M.: Safety verification of deep
  neural networks.
\newblock In: R.~Majumdar, V.~Kun{\v{c}}ak (eds.) Computer Aided Verification,
  pp. 3--29. Springer International Publishing, Cham (2017)

\bibitem{acasxu}
Julian, K.D., Lopez, J., Brush, J.S., Owen, M.P., Kochenderfer, M.J.: Policy
  compression for aircraft collision avoidance systems.
\newblock In: 2016 IEEE/AIAA 35th Digital Avionics Systems Conference (DASC),
  pp. 1--10 (2016).
\newblock \doi{10.1109/DASC.2016.7778091}

\bibitem{reluplex}
Katz, G., Barrett, C., Dill, D.L., Julian, K., Kochenderfer, M.J.: Reluplex: An
  efficient smt solver for verifying deep neural networks.
\newblock In: R.~Majumdar, V.~Kun{\v{c}}ak (eds.) Computer Aided Verification,
  pp. 97--117. Springer International Publishing, Cham (2017)

\bibitem{marabou}
Katz, G., Huang, D.A., Ibeling, D., Julian, K., Lazarus, C., Lim, R., Shah, P.,
  Thakoor, S., Wu, H., Zelji{\'c}, A., et~al.: The marabou framework for
  verification and analysis of deep neural networks.
\newblock In: International Conference on Computer Aided Verification, pp.
  443--452. Springer (2019)

\bibitem{peregrinn}
Khedr, H., Ferlez, J., Shoukry, Y.: Effective formal verification of neural
  networks using the geometry of linear regions.
\newblock arXiv preprint arXiv:2006.10864  (2020)

\bibitem{venus2}
Kouvaros, P., Lomuscio, A.: Towards scalable complete verification of relu
  neural networks via dependency-based branching.
\newblock In: Proceedings of the 30th International Joint Conference on
  Artificial Intelligence ({IJCAI}21) (To Appear, 2021)

\bibitem{nvjl}
Liu, C., Arnon, T., Lazarus, C., Kochenderfer, M.J.: Neuralverification.jl:
  Algorithms for verifying deep neural networks.
\newblock In: ICLR 2019 Debugging Machine Learning Models Workshop (2019).
\newblock
  \urlprefix\url{https://debug-ml-iclr2019.github.io/cameraready/DebugML-19_paper_22.pdf}

\bibitem{lopez2022archcomp_ainncs}
Lopez, D.M., Althoff, M., Benet, L., Chen, X., Fan, J., Forets, M., Huang, C.,
  Johnson, T.T., Ladner, T., Li, W., Schilling, C., Zhu, Q.: Arch-comp22
  category report: Artificial intelligence and neural network control systems
  (ainncs) for continuous and hybrid systems plants.
\newblock In: G.~Frehse, M.~Althoff, E.~Schoitsch, J.~Guiochet (eds.)
  Proceedings of 9th International Workshop on Applied Verification of
  Continuous and Hybrid Systems (ARCH22), \emph{EPiC Series in Computing},
  vol.~90, pp. 142--184. EasyChair (2022).
\newblock \doi{10.29007/wfgr}.
\newblock \urlprefix\url{https://easychair.org/publications/paper/C1J8}

\bibitem{lopez2019archcomp_ainncs}
Lopez, D.M., Musau, P., Tran, H.D., Dutta, S., Carpenter, T.J., Ivanov, R.,
  Johnson, T.T.: Arch-comp19 category report: Artificial intelligence and
  neural network control systems (ainncs) for continuous and hybrid systems
  plants.
\newblock In: G.~Frehse, M.~Althoff (eds.) ARCH19. 6th International Workshop
  on Applied Verification of Continuous and Hybrid Systems, \emph{EPiC Series
  in Computing}, vol.~61, pp. 103--119. EasyChair (2019).
\newblock \doi{10.29007/rgv8}.
\newblock \urlprefix\url{https://easychair.org/publications/paper/BFKs}

\bibitem{oval3}
Lu, J., Kumar, M.P.: Neural network branching for neural network verification.
\newblock In: International Conference on Learning Representations (2020)

\bibitem{MadryMSTV18}
Madry, A., Makelov, A., Schmidt, L., Tsipras, D., Vladu, A.: Towards deep
  learning models resistant to adversarial attacks.
\newblock In: 6th International Conference on Learning Representations, {ICLR}
  2018, Vancouver, BC, Canada, April 30 - May 3, 2018, Conference Track
  Proceedings. OpenReview.net (2018).
\newblock \urlprefix\url{https://openreview.net/forum?id=rJzIBfZAb}

\bibitem{MirmanGV18}
Mirman, M., Gehr, T., Vechev, M.T.: Differentiable abstract interpretation for
  provably robust neural networks.
\newblock In: J.G. Dy, A.~Krause (eds.) Proceedings of the 35th International
  Conference on Machine Learning, {ICML} 2018, Stockholmsm{\"{a}}ssan,
  Stockholm, Sweden, July 10-15, 2018, \emph{Proceedings of Machine Learning
  Research}, vol.~80, pp. 3575--3583. {PMLR} (2018).
\newblock \urlprefix\url{http://proceedings.mlr.press/v80/mirman18b.html}

\bibitem{vnncomp22}
M\"{u}ller, M.N., Brix, C., Bak, S., Liu, C., Johnson, T.T.: The third
  international verification of neural networks competition ({VNN-COMP 2022}):
  Summary and results (2022).
\newblock \doi{10.48550/arXiv.2212.10376}.
\newblock \urlprefix\url{https://arxiv.org/abs/2212.10376}

\bibitem{MullerSABR}
M{\"{u}}ller, M.N., Eckert, F., Fischer, M., Vechev, M.T.: Certified training:
  Small boxes are all you need.
\newblock CoRR \textbf{abs/2210.04871} (2022).
\newblock \doi{10.48550/arXiv.2210.04871}.
\newblock \urlprefix\url{https://doi.org/10.48550/arXiv.2210.04871}

\bibitem{eran5}
M{\"u}ller, M.N., Makarchuk, G., Singh, G., P{\"u}schel, M., Vechev, M.: Prima:
  Precise and general neural network certification via multi-neuron convex
  relaxations.
\newblock arXiv preprint arXiv:2103.03638  (2021)

\bibitem{proofgeneration}
O.~Isac C.~Barrett, M.Z., Katz, G.: Neural network verification with proof
  production.
\newblock In: 22nd International Conference on Formal Methods in Computer-Aided
  Design (FMCAD) (2022)

\bibitem{PalmaIBPR}
Palma, A.D., Bunel, R., Dvijotham, K., Kumar, M.P., Stanforth, R.: {IBP}
  regularization for verified adversarial robustness via branch-and-bound.
\newblock CoRR \textbf{abs/2206.14772} (2022).
\newblock \doi{10.48550/arXiv.2206.14772}.
\newblock \urlprefix\url{https://doi.org/10.48550/arXiv.2206.14772}

\bibitem{eran6}
Serre, F., Müller, C., Singh, G., P{\"u}schel, M., Vechev, M.: Scaling
  polyhedral neural network verification on {GPU}s.
\newblock In: Proc. Machine Learning and Systems (MLSys) (2021)

\bibitem{ShiWZYH21}
Shi, Z., Wang, Y., Zhang, H., Yi, J., Hsieh, C.: Fast certified robust training
  with short warmup.
\newblock In: M.~Ranzato, A.~Beygelzimer, Y.N. Dauphin, P.~Liang, J.W. Vaughan
  (eds.) Advances in Neural Information Processing Systems 34: Annual
  Conference on Neural Information Processing Systems 2021, NeurIPS 2021,
  December 6-14, 2021, virtual, pp. 18335--18349 (2021).
\newblock
  \urlprefix\url{https://proceedings.neurips.cc/paper/2021/hash/988f9153ac4fd966ea302dd9ab9bae15-Abstract.html}

\bibitem{dnnv}
Shriver, D., Elbaum, S., Dwyer, M.B.: Dnnv: A framework for deep neural network
  verification.
\newblock In: A.~Silva, K.R.M. Leino (eds.) Computer Aided Verification, pp.
  137--150. Springer International Publishing, Cham (2021)

\bibitem{dnnf}
Shriver, D., Elbaum, S.G., Dwyer, M.B.: Reducing {DNN} properties to enable
  falsification with adversarial attacks.
\newblock In: 43rd {IEEE/ACM} International Conference on Software Engineering,
  {ICSE} 2021, Madrid, Spain, 22-30 May 2021, pp. 275--287. {IEEE} (2021).
\newblock \doi{10.1109/ICSE43902.2021.00036}.
\newblock \urlprefix\url{https://doi.org/10.1109/ICSE43902.2021.00036}

\bibitem{eran2}
Singh, G., Ganvir, R., P\"{u}schel, M., Vechev, M.: Beyond the single neuron
  convex barrier for neural network certification.
\newblock In: Advances in Neural Information Processing Systems 32, pp.
  15098--15109. Curran Associates, Inc. (2019)

\bibitem{eran4}
Singh, G., Gehr, T., Mirman, M., P\"{u}schel, M., Vechev, M.: Fast and
  effective robustness certification.
\newblock In: S.~Bengio, H.~Wallach, H.~Larochelle, K.~Grauman,
  N.~Cesa-Bianchi, R.~Garnett (eds.) Advances in Neural Information Processing
  Systems 31, pp. 10802--10813. Curran Associates, Inc. (2018).
\newblock
  \urlprefix\url{http://papers.nips.cc/paper/8278-fast-and-effective-robustness-certification.pdf}

\bibitem{eran3}
Singh, G., Gehr, T., P\"{u}schel, M., Vechev, M.: An abstract domain for
  certifying neural networks.
\newblock Proc. ACM Program. Lang. \textbf{3}(POPL), 41:1--41:30 (2019)

\bibitem{eran1}
Singh, G., Gehr, T., Püschel, M., Vechev, M.: Boosting robustness
  certification of neural networks.
\newblock In: Proc. International Conference on Learning Representations (ICLR)
  (2019)

\bibitem{intriguingproperties}
Szegedy, C., Zaremba, W., Sutskever, I., Bruna, J., Erhan, D., Goodfellow,
  I.J., Fergus, R.: Intriguing properties of neural networks.
\newblock In: Y.~Bengio, Y.~LeCun (eds.) 2nd International Conference on
  Learning Representations, {ICLR} 2014, Banff, AB, Canada, April 14-16, 2014,
  Conference Track Proceedings (2014).
\newblock \urlprefix\url{http://arxiv.org/abs/1312.6199}

\bibitem{vnnlib}
Tacchella, A., Pulina, L., Guidotti, D., Demarchi, S.: The verification of
  neural networks library (vnn-lib).
\newblock \url{https://www.vnnlib.org} (2019)

\bibitem{mipverify}
Tjeng, V., Xiao, K.Y., Tedrake, R.: Evaluating robustness of neural networks
  with mixed integer programming.
\newblock In: ICLR (2019)

\bibitem{nnv1}
Tran, H.D., Bak, S., Xiang, W., Johnson, T.T.: Verification of deep
  convolutional neural networks using imagestars.
\newblock In: 32nd International Conference on Computer-Aided Verification
  (CAV). Springer (2020)

\bibitem{nnv4}
Tran, H.D., Musau, P., Lopez, D.M., Yang, X., Nguyen, L.V., Xiang, W., Johnson,
  T.T.: Parallelizable reachability analysis algorithms for feed-forward neural
  networks.
\newblock In: Proceedings of the 7th International Workshop on Formal Methods
  in Software Engineering (FormaliSE'19), FormaliSE '19, pp. 31--40. IEEE
  Press, Piscataway, NJ, USA (2019).
\newblock \doi{10.1109/FormaliSE.2019.00012}

\bibitem{nnv3}
Tran, H.D., Musau, P., Lopez, D.M., Yang, X., Nguyen, L.V., Xiang, W., Johnson,
  T.T.: Star-based reachability analysis for deep neural networks.
\newblock In: 23rd International Symposium on Formal Methods (FM'19). Springer
  International Publishing (2019)

\bibitem{nnv2}
Tran, H.D., Yang, X., Lopez, D.M., Musau, P., Nguyen, L.V., Xiang, W., Bak, S.,
  Johnson, T.T.: {NNV}: The neural network verification tool for deep neural
  networks and learning-enabled cyber-physical systems.
\newblock In: 32nd International Conference on Computer-Aided Verification
  (CAV) (2020)

\bibitem{rpm}
Vincent, J.A., Schwager, M.: Reachable polyhedral marching (rpm): A safety
  verification algorithm for robotic systems with deep neural network
  components (2021)

\bibitem{abcrown4}
Wang, S., Zhang, H., Xu, K., Lin, X., Jana, S., Hsieh, C.J., Kolter, Z.:
  {Beta-CROWN}: Efficient bound propagation with per-neuron split constraints
  for complete and incomplete neural network verification.
\newblock arXiv preprint arXiv:2103.06624  (2021)

\bibitem{nnv5}
{Xiang}, W., {Tran}, H., {Johnson}, T.T.: Output reachable set estimation and
  verification for multilayer neural networks.
\newblock IEEE Transactions on Neural Networks and Learning Systems
  \textbf{29}(11), 5777--5783 (2018)

\bibitem{abcrown2}
Xu, K., Shi, Z., Zhang, H., Wang, Y., Chang, K.W., Huang, M., Kailkhura, B.,
  Lin, X., Hsieh, C.J.: Automatic perturbation analysis for scalable certified
  robustness and beyond.
\newblock Advances in Neural Information Processing Systems \textbf{33} (2020)

\bibitem{abcrown3}
Xu, K., Zhang, H., Wang, S., Wang, Y., Jana, S., Lin, X., Hsieh, C.J.: {Fast
  and Complete}: Enabling complete neural network verification with rapid and
  massively parallel incomplete verifiers.
\newblock In: International Conference on Learning Representations (2021).
\newblock \urlprefix\url{https://openreview.net/forum?id=nVZtXBI6LNn}

\bibitem{metasat}
Xu, L., Hutter, F., Hoos, H.H., Leyton-Brown, K.: Satzilla: Portfolio-based
  algorithm selection for sat.
\newblock J. Artif. Int. Res. \textbf{32}(1), 565–606 (2008)

\bibitem{ZhangCXGSLBH20}
Zhang, H., Chen, H., Xiao, C., Gowal, S., Stanforth, R., Li, B., Boning, D.S.,
  Hsieh, C.: Towards stable and efficient training of verifiably robust neural
  networks.
\newblock In: 8th International Conference on Learning Representations, {ICLR}
  2020, Addis Ababa, Ethiopia, April 26-30, 2020. OpenReview.net (2020).
\newblock \urlprefix\url{https://openreview.net/forum?id=Skxuk1rFwB}

\bibitem{ZhangGCP}
Zhang, H., Wang, S., Xu, K., Li, L., Li, B., Jana, S., Hsieh, C., Kolter, J.Z.:
  General cutting planes for bound-propagation-based neural network
  verification.
\newblock CoRR \textbf{abs/2208.05740} (2022).
\newblock \doi{10.48550/arXiv.2208.05740}.
\newblock \urlprefix\url{https://doi.org/10.48550/arXiv.2208.05740}

\bibitem{ZhangWCHD18}
Zhang, H., Weng, T., Chen, P., Hsieh, C., Daniel, L.: Efficient neural network
  robustness certification with general activation functions.
\newblock In: S.~Bengio, H.M. Wallach, H.~Larochelle, K.~Grauman,
  N.~Cesa{-}Bianchi, R.~Garnett (eds.) Advances in Neural Information
  Processing Systems 31: Annual Conference on Neural Information Processing
  Systems 2018, NeurIPS 2018, December 3-8, 2018, Montr{\'{e}}al, Canada, pp.
  4944--4953 (2018).
\newblock
  \urlprefix\url{https://proceedings.neurips.cc/paper/2018/hash/d04863f100d59b3eb688a11f95b0ae60-Abstract.html}

\bibitem{abcrown1}
Zhang, H., Weng, T.W., Chen, P.Y., Hsieh, C.J., Daniel, L.: Efficient neural
  network robustness certification with general activation functions.
\newblock Advances in Neural Information Processing Systems \textbf{31},
  4939--4948 (2018).
\newblock \urlprefix\url{https://arxiv.org/pdf/1811.00866.pdf}

\bibitem{maxpooling}
Zhou, Chellappa: Computation of optical flow using a neural network.
\newblock In: IEEE 1988 International Conference on Neural Networks, pp. 71--78
  vol.2 (1988).
\newblock \doi{10.1109/ICNN.1988.23914}

\end{thebibliography}

\end{document}